\documentclass{article}

\usepackage{microtype}
\usepackage{graphicx}
\usepackage{subfigure}
\usepackage{enumitem}
\usepackage{booktabs} 

\usepackage{hyperref}
\usepackage{multirow}
\usepackage{amssymb}

\usepackage{color,colortbl}
\usepackage[dvipsnames]{xcolor}

\definecolor{color2}{HTML}{009B55}
\definecolor{color3}{HTML}{7d7c84}
\definecolor{mygray}{HTML}{F0F0F0}
\definecolor{azureblue}{rgb}{0,0.5,1}

\usepackage[accepted]{icml2024}

\usepackage{amsmath}
\usepackage{amssymb}
\usepackage{mathtools}
\usepackage{amsthm}

\usepackage[capitalize,noabbrev]{cleveref}

\theoremstyle{plain}

\theoremstyle{definition}

\theoremstyle{remark}

\newcommand{\graph}[1]{MAG}
\newcommand{\model}[1]{\textsc{MAGDi}}
\newcommand{\method}[0]{\model{}}
\newcommand{\sparagraph}[1]{\textbf{#1 \hspace{0.2em}}}

\newcommand{\mytitle}[1]{\model{}: Structured Distillation of Multi-Agent Interaction Graphs \\Improves Reasoning in Smaller Language Models}

\usepackage[textsize=tiny]{todonotes}

\icmltitlerunning{\textsc{MADGi}: Structured Distillation of Multi-Agent Interaction Graphs}
\begin{document}

\twocolumn[
\icmltitle{\mytitle{}}
\icmlsetsymbol{equal}{*}

\begin{icmlauthorlist}
\icmlauthor{Justin Chih-Yao Chen}{equal,yyy}
\icmlauthor{Swarnadeep Saha}{equal,yyy}
\icmlauthor{Elias Stengel-Eskin}{yyy}
\icmlauthor{Mohit Bansal}{yyy}

\end{icmlauthorlist}

\icmlaffiliation{yyy}{UNC Chapel Hill} 

\icmlcorrespondingauthor{Justin Chih-Yao Chen}{cyaochen@cs.unc.edu}

\icmlkeywords{}

\vskip 0.3in
]

\printAffiliationsAndNotice{\icmlEqualContribution} 
\begin{abstract}
Multi-agent interactions between Large Language Model (LLM) agents have shown major improvements on diverse reasoning tasks. However, these involve long generations from multiple models across several rounds, making them expensive. 
Moreover, these multi-agent approaches fail to provide a final, single model for efficient inference.
To address this, we introduce \method{}, a new method for \emph{structured distillation of the reasoning interactions between multiple LLMs into smaller LMs}.
\model{} teaches smaller models by representing multi-agent interactions as graphs,
augmenting a base student model with a graph encoder, and distilling knowledge using three objective functions: next-token prediction, a contrastive loss between correct and incorrect reasoning, and a graph-based objective to model the interaction structure. 
Experiments on seven widely-used commonsense and math reasoning benchmarks show that 
\method{}
improves the reasoning capabilities of smaller models, outperforming several methods that distill from a single teacher and multiple teachers. Moreover, \method{} also demonstrates an order of magnitude higher efficiency over its teachers.
We conduct extensive analyses to show that \method{} (1) enhances the generalizability to out-of-domain tasks, (2) scales positively with the size and strength of the base student model, and (3) obtains larger improvements (via our multi-teacher training) when applying self-consistency -- an inference technique that relies on model diversity.\footnote{Code/data: \url{https://github.com/dinobby/MAGDi}.}
\end{abstract}

\begin{figure}
    \centering
    \includegraphics[width=\linewidth]{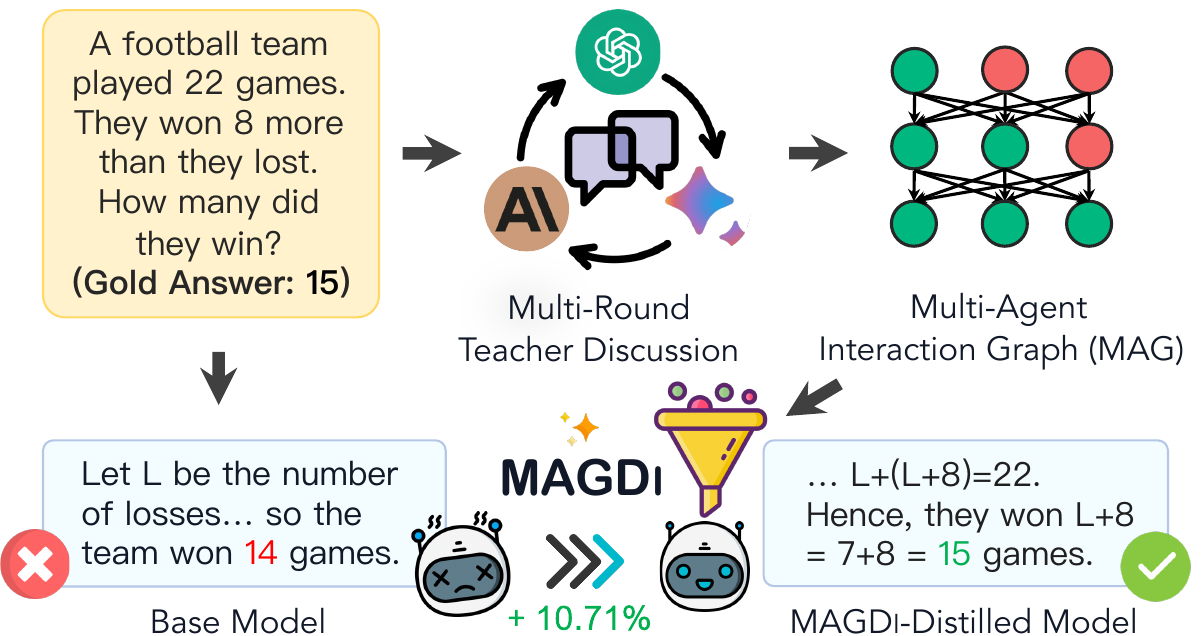}
    \vspace{-1.5em}
    \caption{Overview of our distillation method. Given a reasoning problem, multiple teacher-LLMs engage in a multi-round discussion, leading to the generation of a multi-agent interaction graph (\graph{}). Then our structured distillation method, \model{} distills reasoning knowledge from these graphs into a base student model.  
    }
    \vspace{-1em}
    \label{fig:teaser}
\end{figure}

\section{Introduction} \label{sec:intro}

Debate and dialogue are natural ways to improve reasoning: we form our best ideas not in isolation, but by refining and discussing them with others. 
Similarly, we can improve Large Language Models (LLMs) -- which often exhibit impressive multi-step reasoning capabilities~\citep{wei2023chainofthought, kojima2022large} -- by allowing multiple LLM instances to interact in a discussion~\citep{du2023improving, chen2023reconcile, wu2023autogen}. 
These interactive frameworks enable each agent to iteratively refine its reasoning by obtaining feedback from others, thereby leading to a better consensus at the end of multiple interaction rounds.

\begin{figure*}
\begin{minipage}[b]{.63\textwidth}
\subfigure[]{\label{fig:intro_a}\includegraphics[width=1.0\textwidth]{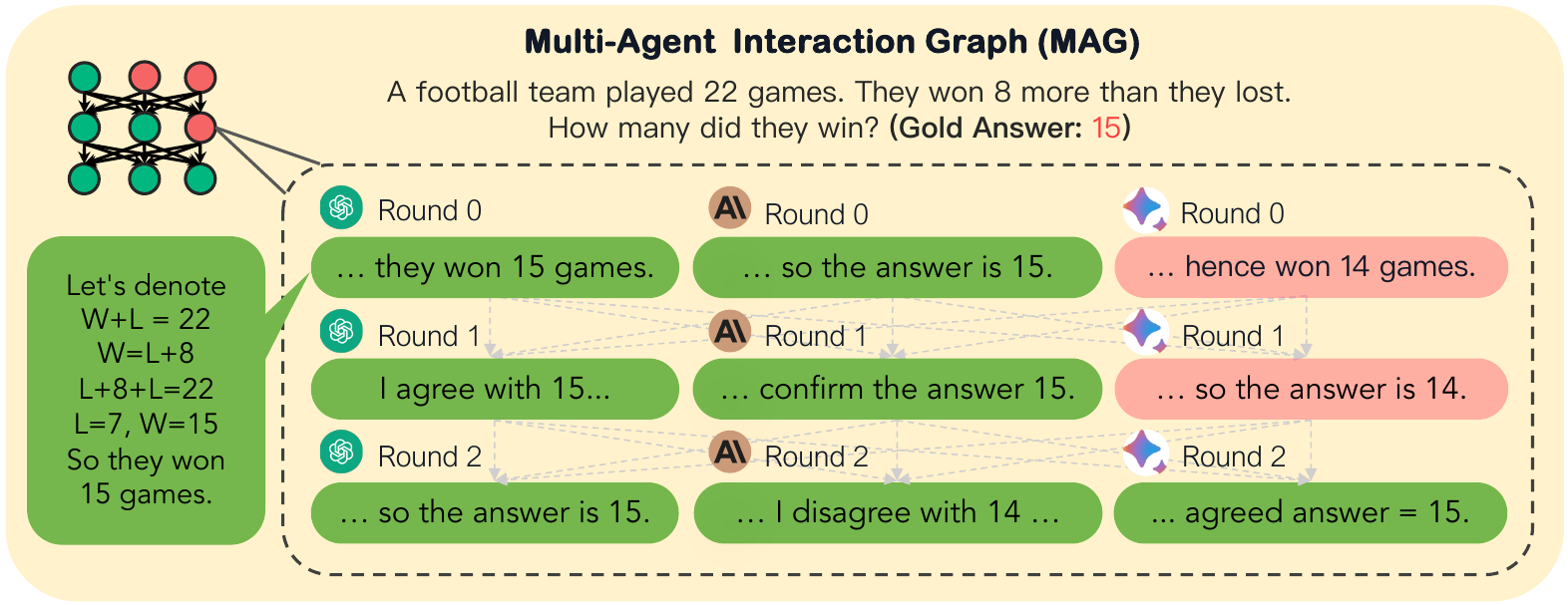}}
\end{minipage}
\begin{minipage}[b]{1\textwidth}
\begin{minipage}[b]{1\textwidth}
    \begin{minipage}[b]{.18\textwidth}
    \subfigure[]{\label{fig:intro_b}\includegraphics[width=\textwidth]{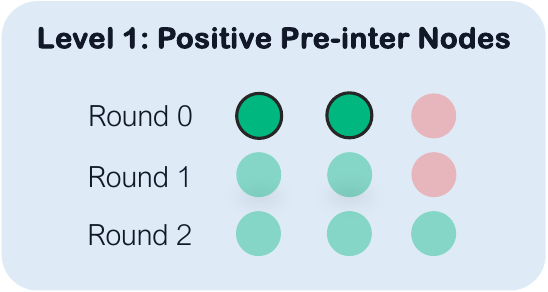}}
    \end{minipage}
    \begin{minipage}[b]{.18\textwidth}
    \subfigure[]{\label{fig:intro_d}\includegraphics[width=\textwidth]{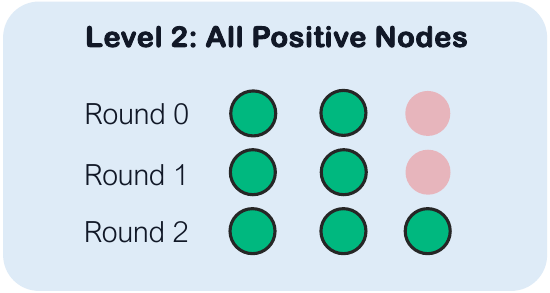}}
    \end{minipage}
\end{minipage}
\begin{minipage}[b]{1\textwidth} 
    \begin{minipage}[b]{.18\textwidth}
    \subfigure[]{\label{fig:intro_c}\includegraphics[width=\textwidth]{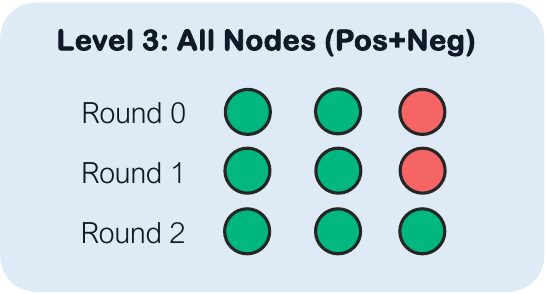}}
    \end{minipage}
    \begin{minipage}[b]{.18\textwidth}
    \subfigure[]{\label{fig:intro_e}\includegraphics[width=\textwidth]{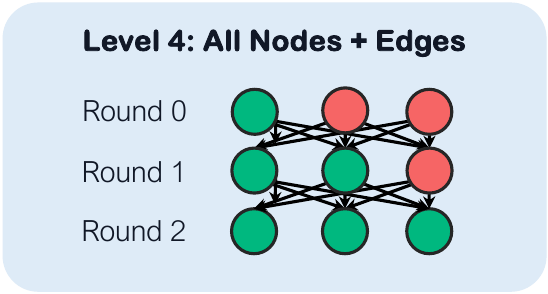}}
    \end{minipage}
\end{minipage}
\end{minipage}
    \vspace{-1.2em}
    \caption{\textbf{Left (a):} Illustration of a Multi-Agent Interaction Graph (\graph{}) constructed with GPT4, Bard, and Claude2 collaboratively solving a math reasoning problem over three discussion rounds. 
    \textbf{Right (b-e):} The four levels that characterize our structured distillation method (\model{}); each level progressively distills knowledge from the highlighted components of a \graph{}.
    \vspace{-1em}
    }
    \label{fig:intro}
\end{figure*}

Discussion frameworks are typically built on top of proprietary models, e.g., GPT-4, Bard, Claude, etc., which can act as general conversational agents, handle long contexts, and follow instructions~\citep{bubeck2023sparks}. 
However, these models are also computationally and monetarily expensive, especially when used in multi-round interactions, which require numerous long-token length inference calls to the underlying LLMs.\footnote{
For example, one such multi-LLM interaction method, ReConcile \citep{chen2023reconcile} uses around 1900 tokens per sample on a math reasoning task, with other discussion-based methods (e.g., \citet{du2023improving, wu2023autogen}) using even larger token budgets.}
Moreover, these frameworks do not result in a final, joint model that can then be directly used for inference and instead require invoking all interacting LLMs at test time.
To reduce this cost and train a small, affordable yet capable model, we tackle the problem of teaching reasoning to smaller language models via \emph{structured distillation of the interactions between multiple stronger teacher models}.
Specifically, we develop a structured distillation method, \textbf{M}ulti-\textbf{A}gent Interaction \textbf{G}raphs \textbf{Di}stillation, or \model{}, that enables a student model to learn from multi-teacher interactions, with the goal of developing a performant and efficient standalone alternative to expensive multi-agent setups. On seven popular benchmarks in both commonsense and math reasoning, we find increasing improvements over distillation baselines as we incorporate more levels of teacher interactions and structure.

Multi-agent, multi-round interactions are characterized by their participating agents, the number of interaction rounds, and an interaction function defining what information an agent has access to while generating its responses.
This function gives rise to a structure between agents and rounds. To learn from this structure, we represent it in   
\textbf{M}ulti-\textbf{A}gent Interaction \textbf{G}raphs (\graph{}), 
a graph-based encoding of multi-agent interactions (\S\ref{sec:migs_def}). 
See Fig.~\ref{fig:intro_a} for an example. Concretely, a \graph{} is a directed acyclic graph (DAG) wherein each node represents an agent's generation (in this case, the Chain-of-Thought reasoning~\citep{wei2023chainofthought} for a given problem) in a discussion round, annotated with a binary label indicating whether the answer is correct.
The edges denote the discussion's structure, indicating which previous turns agents are responding to. 
\graph{}s are an intuitive and generalizable way of representing the levels of many multi-agent interactions with varying conversation patterns~\citep{wu2023autogen}, and will allow us to distill this information into a student model for performing zero-shot inference from just the question (i.e., \graph{}s are not required at test-time). 

\looseness-1
Given a reasoning problem, \graph{}s capture rich knowledge of (1) diverse \emph{pre-} and \emph{post-interaction correct reasoning chains} generated by different LLMs (green nodes in Fig.~\ref{fig:intro_a}), (2) diverse and challenging \emph{incorrect reasoning chains} generated by different LLMs that are refined over interaction rounds (red nodes in Fig.~\ref{fig:intro_a}), and (3) an \emph{iterative and structured (graph-based) interaction process} that enables this refinement of model reasoning (edges in Fig.~\ref{fig:intro_a}). 
We capture all this knowledge via the following four levels of \graph{} components, which are then used in our distillation method, \model{} (\S\ref{sec:structured}), and further tested as part of our experiments (\S\ref{sec:main_results}).

\begin{itemize}[topsep=0pt,itemsep=0pt, labelwidth=0pt, labelindent=0pt, leftmargin=0pt,label={}]
\item \textbf{Level 1: Learning from multiple teachers.} The student learns from the correct reasoning of \emph{multiple} teachers, rather than one (correct pre-interaction nodes in a \graph{}, Fig.~\ref{fig:intro_b}).
\item \textbf{Level 2: Learning from teacher interactions.} The student learns from both pre- and \emph{post-interaction} data between multiple teachers (all correct nodes in a \graph{}, Fig.~\ref{fig:intro_d}).
\item \textbf{Level 3: Learning from negative reasoning.} The student additionally distills from \emph{negative or incorrect} reasoning from the teacher models (all nodes in a \graph{}, Fig.~\ref{fig:intro_c}).
\item \textbf{Level 4: Learning from structure.} The student learns from the output and \emph{graph-structure} of teacher LLM interactions (all nodes and edges in a \graph{}, Fig.~\ref{fig:intro_e}).
\end{itemize}

Note that each level builds on the prior levels, motivating our main research question:

\emph{\textbf{Research Question:} How can we effectively distill from diverse teacher interactions into a smaller, efficient student model across increasing levels of interaction structure, also demonstrating scalability and generalizability?}

These levels also shape \method{}, our structured distillation method. 
\method{} enables a student model to learn from our graph-structured interaction data (\graph{}s), with the goal of developing a performant and efficient standalone alternative to expensive multi-agent setups. 
We first construct a training dataset of MAGs from a high-performing multi-agent discussion framework~\citep{chen2023reconcile}, featuring discussions between three API-based LLMs: GPT-4, Bard, and Claude2 (\S\ref{sec:training_data}).
We then develop student models augmented with a Graph Neural Network (GNN) for learning \emph{structure-aware} representations of positive (correct) and negative (incorrect) reasoning chains and fine-tune them on \graph{} data.
\method{}'s three fine-tuning objectives are aligned to the four levels: (1) next-token prediction (Levels 1-2), (2) a contrastive loss between correct and incorrect reasoning (Level 3), and (3) a graph-based node classification loss (Level 4). 
These objectives capture all useful signals in MAGs (i.e., teachers' \emph{correct} and \emph{incorrect} reasoning as well as their underlying \emph{conversation structure}). At test time, the distilled model performs zero-shot inference given just the \emph{question} and the \emph{base model} (without the GNN).

We evaluate \method{}'s effectiveness on seven widely-used commonsense (StrategyQA, CommonsenseQA, ARC-c, BoolQ) and math (GSM8K, MATH, SVAMP) reasoning benchmarks, consistently establishing the following findings across datasets and domains:
\begin{itemize}[noitemsep, topsep=0pt, wide=0pt, leftmargin=*]
\item \textbf{Multi-teacher distillation improves student performance.} When compared directly to distilling from a single teacher, distilling from \emph{multiple teachers} improves performance (Level 1).
\item \textbf{The value of teacher interactions}: Distilling from the \emph{post-interaction} outputs of teachers further improves students (Level 2). 
\item \textbf{Negative reasoning helps.} Adding a contrastive objective to learn from \emph{incorrect} reasoning provides a valuable signal to the student model (Level 3).
\item \textbf{Distilling from structure maximizes accuracy.} When \model{} distills from the first 3 levels \emph{and the structure} of a MAG, the student achieves the highest accuracy, e.g., up to $10\%$ absolute improvement over a zero-shot baseline and up to $4\%$ over the best single-teacher baseline.
\item \textbf{\method{} balances performance with efficiency.} \method{}-distilled models reduce the number of tokens predicted at test time by up to $9$x while outperforming all single-teacher distillation baselines. 
\end{itemize}

Building on these results, we further analyze \method{} along the following axes:
\begin{itemize}[noitemsep, topsep=0pt, wide=0pt, leftmargin=*]
\item \textbf{Generalizability.} \method{} can be used to produce a unified joint multi-task learning model that performs well on multiple domains at once and also generalizes well to held-out datasets not seen during training.
\item \textbf{Scalability.} \method{} scales positively with the size and strength of the base student model.
\item \textbf{Diversity.} The output diversity resulting from our multi-teacher training improves self-consistency \citep{wang.y.2022self}, an inference-time ensemble method relying on diverse model answers. 
\end{itemize}

\begin{figure*}
    \centering
    \includegraphics[width=\textwidth]{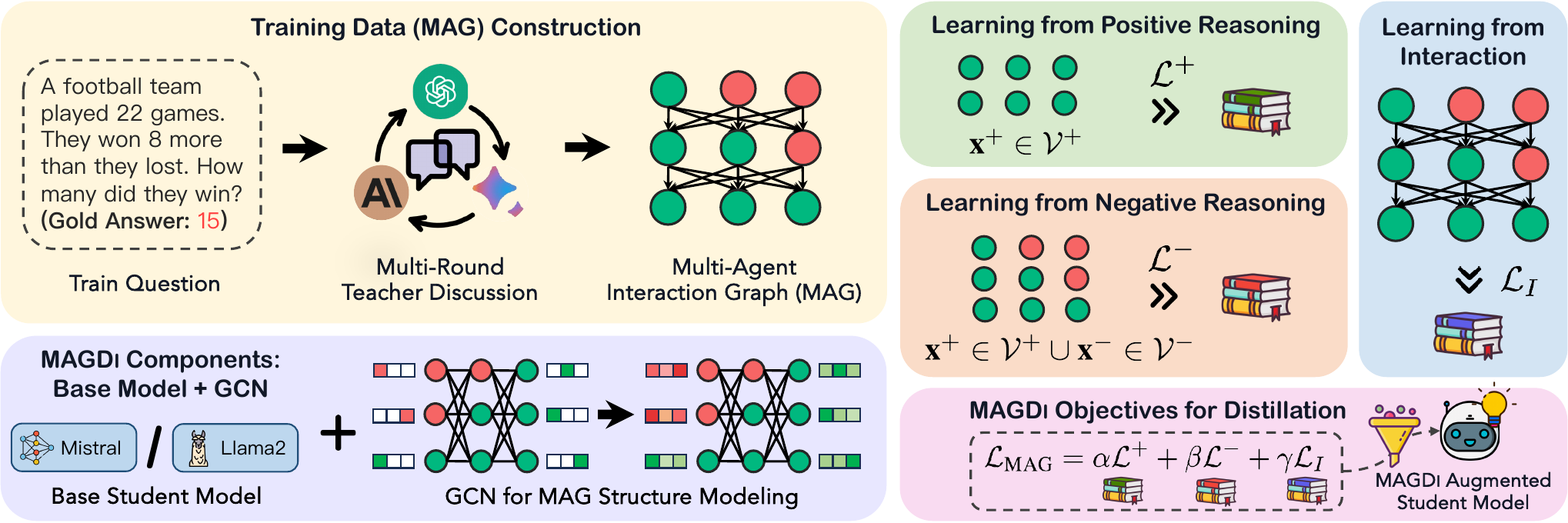}
    \vspace{-15pt}
    \caption{
    \textbf{Training Data Construction:} Given a reasoning problem, multiple teachers go through a multi-round discussion process, generating multi-agent interaction graphs (\graph{}s). \textbf{\model{}:} Our structured distillation method augments a base student model with a Graph Neural Network (specifically, a GCN) to learn structure-aware representations of reasoning chains. The resultant model is then fine-tuned with a combination of three objectives involving positive chains, negative chains, and the underlying interactions.
    }
    \vspace{-1em}
    \label{fig:overview}
\end{figure*}

\section{Related Work}

\sparagraph{Knowledge Distillation.} Knowledge distillation has proven effective in transferring knowledge from a larger teacher model to a more compact student model~\cite{hinton2015distilling, bucilua.c.2006model, chen.t.2020big} including distillation from multiple teacher models~\citep{you2017learning, yang2020model}. 
Following recent work, we focus on distillation from \emph{samples} from a model distribution, or symbolic distillation \citep{west-etal-2022-symbolic}, a form of distillation especially common on LLMs, e.g. in instruction tuning~\citep{wang.y.2022self, taori.r.2023stanford, vicuna2023}, where instruction-question-answer triples are sourced from a teacher model. 
In reasoning with LLMs specifically, recent work has distilled reasoning knowledge from a \emph{single} larger teacher model to a smaller student model~\cite{magister2022teaching, shridhar2023distilling, fu2023specializing, ho2022large, saha2023can, mukherjee2023orca, mitra2023orca, li-etal-2023-symbolic, deng2023implicit, liu.b.2023tinygsm} using Chain-of-Thought prompting~\citep[CoT;][]{wei2023chainofthought} and also, a combination of multiple prompting techniques~\citep{chenglin.l.2023mixed, mukherjee2023orca, mitra2023orca}.
Past work has also distilled modular trajectories from a GPT-4 teacher for solving interactive tasks~\citep{chen2023fireact, yin2023lumos}. 
Overall, different from these single-teacher settings, we delve into the realm of knowledge distillation from \emph{multiple teachers}. 
Going one step further, we learn from the \emph{interactions} between teachers, bringing in fresh challenges on flexibly representing and modeling these interactions.

\sparagraph{Graph-based Interactions.} Dialogues, debates, and multi-party conversations~\citep{kirchhoff2003directions, leifeld2018discourse, wei2023multi} have a long, rich history of being modeled as graphs for different downstream tasks such as emotion and sentiment identification~\citep{ghosal2019dialoguegcn, shen2021directed}, dialogue act recognition~\citep{qin2021co}, dialogue summarization~\citep{chen2021structure}, and machine reading~\citep{ouyang2021dialogue}. 
Our motivation in this work differs in two major respects. Firstly, we focus on model-model interactions rather than the human-human or human-model interactions from past work.
Secondly, our objective is to enhance reasoning capabilities in smaller student models through structured distillation, as opposed to utilizing graph modeling for downstream graph tasks.

\section{Method}
\looseness-1
In Sec.~\ref{sec:migs_def}, we provide a general, formal description of \graph{}, our graph-based representation of multi-agent interactions. Sec.~\ref{sec:training_data} then describes the construction of \graph{}s for several tasks that will serve as training data for distillation.
In Sec.~\ref{sec:data_analysis}, we analyze this training data in terms of its structural properties. Lastly, in Sec.~\ref{sec:structured}, we describe \model{}, our structured distillation method for learning from \graph{}s. 

\subsection{Multi-Agent Interaction Graph (\graph{})} \label{sec:migs_def}
\sparagraph{Definition.} Consider a collaborative multi-agent setting, where $\mathcal{A} = \{A_i\}_{i=1}^n$ agents are interacting with each other for $r$ rounds to solve a task. A multi-agent interaction graph (\graph{}) is a structured encoding of the interactions between these $\mathcal{A}$ agents over $r$ rounds. Formally, a \graph{} is a Directed Acyclic Graph $\mathcal{G} = (\mathcal{V}, \mathcal{E})$, where $\mathcal{V}$ is a set of nodes and $\mathcal{E}$ is a set of directed edges. A node $v_{i,j} \in \mathcal{V}$ represents the output of an agent $A_i \in \mathcal{A}$ in interaction round $j \in [0,r]$.\footnote{Round 0 refers to an agent's pre-interaction output.} For example, see Fig.~\ref{fig:intro_a} where each node represents an agent's Chain-of-Thought reasoning to the question. Edges denote conditional dependencies between the agents' interactions and encode how an agent's output is refined over interaction rounds. Specifically, we define a directed edge between two nodes if the target node's generation is conditioned on the source node's generation.

\sparagraph{Example of a \graph{}.} \graph{}s can be generally defined for arbitrary agents and interaction patterns. 
We focus on reasoning problems as a general class of domains where interaction has positive impacts, defining MAGs for LLM agents on commonsense and math reasoning tasks.
Past works have defined several such interaction frameworks~\cite{du2023improving, liang2023encouraging, chen2023reconcile} for which \graph{}s can be defined. Of these, we choose \textsc{ReConcile}~\citep{chen2023reconcile} as our interaction framework because of (1) its performance: it obtains the highest performance across multiple benchmarks, and (2) its flexibility: it allows each agent to converse in natural language following the generic Chain-of-Thought reasoning paradigm and thus, can be readily applied to any downstream task where CoT is applicable. \graph{}s make minimal assumptions about the contents of nodes and hence, can be similarly defined for agents conversing using other prompting techniques~\citep{chen2022program}.   Fig.~\ref{fig:intro_a} shows an example of a \textsc{ReConcile}-based \graph{} for a math problem. Since the nodes in a \graph{} represent model responses (either correct or incorrect), we additionally annotate each node $v \in \mathcal{V}$ with a binary label $y_v \in \{0,1\}$, indicating the correctness of the answer. These are marked with green and red circles in Fig.~\ref{fig:intro}. We refer to these two sets of nodes as $\mathcal{V}^+$ and $\mathcal{V}^-$ respectively such that $\mathcal{V} = \mathcal{V}^+ \bigcup \mathcal{V}^-$. Following the interaction pattern of \textsc{ReConcile} that conditions each subsequent round of interaction on \emph{all} agent outputs from the previous round, 
we define edges from all source nodes $v_{i,j}$ to all target nodes $v_{k, j+1}$, i.e.,
$(v_{i,j}, v_{k, j+1}) \in \mathcal{E}~\forall i,k \in [1,n], \forall j \in [0,r]$.

\subsection{Training Data (\graph{}) Construction}
\label{sec:training_data}

With the specifics of a \graph{} defined, we now want to construct these graphs for a given task. These would then serve as data for training distilled models. Given a reasoning problem (e.g., a math word problem), we follow \textsc{ReConcile} 
and use GPT-4, Bard, and Claude2 as the three interacting LLM agents for a maximum of three rounds (see \citet{chen2023reconcile} for further details of the framework). 
The discussion continues until a consensus is reached, i.e., all agents agree on the same answer.
This means that when there is no interaction (i.e., all agents' initial answers are the same), a \graph{} will have 3 disconnected nodes, one for each agent (and no edges). When there is a single round of interaction, it will have 6 nodes and 9 edges, and so on.  In summary, for a given task, our training data will consist of graphs that can be grouped into four structural types, based on the number of interaction rounds. We will refer to these graph types as $\mathcal{G}_0$, $\mathcal{G}_1$, $\mathcal{G}_2$, and $\mathcal{G}_3$ (with the subscript denoting the number of rounds). Fig.~\ref{fig:intro_a} is an example of a $\mathcal{G}_2$ graph. 

\sparagraph{Benchmarks.} Following the above framework, we construct \graph{}s for 5 widely-used benchmarks on commonsense and math reasoning: (1) StrategyQA~\cite{geva-etal-2021-aristotle}, (2) CommonsenseQA~\cite{talmor-etal-2019-commonsenseqa, aggarwaletal2021ecqa}, (3) AI2 Reasoning Challenge~\cite{clark2018think}, (4) GSM8K~\cite{cobbe2021gsm8k} and (5) MATH~\cite{hendrycksmath2021}. 
We also experiment with 2 OOD datasets: BoolQ~\citep{clark-etal-2019-boolq} and SVAMP~\citep{patel-etal-2021-nlp} for which we \emph{do not} construct \graph{}s and are exclusively used to test the transfer of our model.

\subsection{Statistics of Training Data (\graph{}s)}
\label{sec:data_analysis}

We construct $1000$ training \graph{}s for each in-domain benchmark. We categorize the data along three dimensions: rounds, agents, and graph structures. See Appendix~\cref{tab:dataset} for statistics along each dimension and for each benchmark.

\noindent \textbf{Round.} Recall that a \graph{} consists of nodes belonging to different interaction rounds $i \in [0,3]$. All datasets have more nodes in lower rounds, indicating that consensus between agents is typically achieved in these earlier rounds. 
The number of nodes in the later rounds is also an indicator of the difficulty of the benchmark. For example, MATH has the most number of round-3 nodes, suggesting that even after two rounds of discussion, the strong teacher LLMs do not converge on a single answer.

\noindent \textbf{Agent.} We can also group the \graph{} nodes based on the agent generating the response at that node (GPT4/Claude2/Bard). The number of nodes for each agent is the same because all agents engage in all rounds.

\noindent \textbf{Graph Structure.} Lastly, we also show the break-down of different \graph{} structures. Like nodes, $\mathcal{G}_0$ graphs are the most represented in our datasets while $\mathcal{G}_3$ graphs are the least, and all graph structures add up to 1K data points per task. We show examples of \graph{}s in Appendix~\ref{appendix:qual}. Using these \graph{}s as supervision, we train task-specific and multi-task distilled models, as discussed below.

\subsection{\model{}: Structured Distillation from \graph{}s} \label{sec:structured}
We now discuss our proposed structured distillation method, \model{}, that distills reasoning capabilities into a smaller student model via multi-teacher interaction graphs (\graph{}s).

\sparagraph{\model{} Overview.} Broadly, \model{} performs structured distillation by augmenting a student model (e.g., Mistral-7B-Instruct or LLaMA-2-7B-Chat) with a light-weight Graph Neural Network (GNN) that is responsible for modeling the `structure' in structured distillation (see Fig.~\ref{fig:overview} `\model{} Components'). This augmented student model is then fine-tuned with a combination of three objectives that distill knowledge from the positive nodes, negative nodes, and the edges (i.e., interactions) in a \graph{} (see Fig.~\ref{fig:overview} `\model{} Objectives for Distillaton'). We denote the base model as $p_\theta(\cdot)$ and the input reasoning problem as $q$. For brevity, we denote the generation at any \graph{} node as a variable-length sequence of tokens: $\textbf{x} = \{x_1, x_2, ..., x_{|\textbf{x}|}\}$. Below, we describe the three objectives for structurally distilling student LMs.

\sparagraph{Objective 1: Learning from Positive Reasoning.} Learning from a teacher LLM's \textit{correct} reasoning has been shown to improve smaller models~\citep{magister2022teaching, li-etal-2023-symbolic}. Hence, \model{} first fine-tunes the student model on all \emph{correct} reasoning chains $\textbf{x}^+ \in \mathcal{V}^+$ using a standard next-token prediction objective (see Fig.~\ref{fig:overview} `Learning from Positive Reasoning'). The loss is defined as follows.
\begin{equation}
\label{eqn:pos}
\mathcal{L}^+ = - \sum_{\textbf{x}^+ \in \mathcal{V}^+} \sum_{i=1}^{|\textbf{x}^+|} \log p_\theta(x^+_i|\textbf{x}^+_{<i}, q)
\end{equation}
\sparagraph{Objective 2: Learning from Negative Reasoning.}  Teacher LLMs also make mistakes, particularly when solving more challenging problems. However, instead of discarding these, \model{} treats them as \emph{challenging} negatives (generated by a strong teacher) that a student model can learn from by contrasting with positive chains. 
Given a positive reasoning chain $\textbf{x}^+ \in \mathcal{V}^+$ and a negative reasoning chain $\textbf{x}^- \in \mathcal{V}^-$, we first extract representations of these chains by performing a weighted average pool over the final layer's hidden representations of the constituent tokens. We represent these as $h_{\textbf{x}^+} \in \mathbb{R}^{d}$ and $h_{\textbf{x}^-} \in \mathbb{R}^{d}$ respectively where $d$ is the embedding dimension. Using a projection matrix and a $\tanh$ activation, we further project these embeddings to two scalar scores  $s_{\textbf{x}^+} \in [-1, 1]$ and $s_{\textbf{x}^-} \in [-1, 1]$. \model{} then optimizes for the following margin-based objective~\cite{cortes1995support} for pairs of positive and negative chains \{$\textbf{x}^+, \textbf{x}^-$\} $\in \mathcal{V}^+ \times \mathcal{V}^-$ in a \graph{} (see Fig.~\ref{fig:overview} `Learning from Negative Reasoning').
\begin{equation}
\label{eqn:neg}
\mathcal{L}^- = \sum_{\textbf{x}^+ \in \mathcal{V}^+} \sum_{\textbf{x}^- \in \mathcal{V}^-} \max(0, \rho - s_{\textbf{x}^+} + s_{\textbf{x}^-})
\end{equation}
where $\rho \in [-1,1]$ is the margin (set to 1 in our experiments).

\sparagraph{Objective 3: Learning from Interaction.} Beyond distillation from the correct and incorrect nodes, \model{} is also intended to distill from the entire conversational \emph{structure} present in the teachers' discussion. This would allow the student model to summarize the discussion process and acquire knowledge of how a teacher refines its reasoning chain in each discussion round. Hence, while the previous two objectives assumed nodes to be a disconnected set, \model{} removes this assumption by also modeling the edges. 
\model{} achieves this by augmenting the base student model with a Graph Convolution Network (GCN) module~\cite{kipf2017semisupervised}. The goal of the GCN is to learn improved, `structure-aware' representations of reasoning chains (nodes) such that the student model learns to discriminate between correct and incorrect nodes in a \graph{}, eventually leading to better generation of reasoning chains.

Given any positive or negative node $\textbf{x} \in \mathcal{V}$, \model{} generates a node representation $h_{\textbf{x}} \in \mathbb{R}^d$ from the base LM. It then learns structure-aware representations of these nodes with a two-layer GCN 
using the following equation: 
\[h^{(l+1)}_{\textbf{x}} = \sigma(D^{-1} M  h^{(l)}_{\textbf{x}} W^{(l)}) \]
where $M \in |\mathcal{V}| \times |\mathcal{V}|$ is the adjacency matrix with self-connections, $D$ is the diagonal degree matrix of $M$, $\sigma$ is the ReLU activation function, $h^{(l)}_{\textbf{x}}$ and $h^{(l+1)}_{\textbf{x}}$ are the input and updated node representations of the $l$-th layer. We set $h^{(0)}_{\textbf{x}} = h_{\textbf{x}}$ and $W^{(l)}$ as the weight matrix of the $l$-th layer. After two layers of message passing, we obtain the final node representation $h^{(L)}_{\textbf{x}}$, which is now conditioned on the graph structure. \model{} then projects $h^{(L)}_{\textbf{x}}$ with a linear layer parameterized by $W_c \in \mathbb{R}^{d \times C}$ (where $C$ is the number of node labels) 
and applies the softmax function to derive the probability distribution over the node labels $\hat{y}_\textbf{x} = \text{softmax}(h^{(L)}_{\textbf{x}} W_c)$.
Finally, we use cross-entropy loss for the (correct/incorrect) node classification objective over all nodes in a \graph{} (see Fig.~\ref{fig:overview} `Learning from Interaction'),
\begin{equation}
\label{eqn:int}
    \mathcal{L}_{I} = -\sum_{\textbf{x} \in \mathcal{V}} \sum_{i=1}^{C} y_\textbf{x}^{(i)} \log(\hat{y}_{\textbf{x}}^{(i)})
\end{equation}
where $y_\textbf{x}$ is a one-hot encoding of the label of a node $\textbf{x}$.

\sparagraph{Final Objective.} Our final loss, $\mathcal{L}_{\mathit{MAG}}$, as defined below, is a weighted combination of the three losses.
\begin{equation}
\label{eqn:all}
\mathcal{L}_{\mathit{MAG}} = \alpha \mathcal{L}^+ + \beta \mathcal{L}^- + \gamma \mathcal{L}_I
\end{equation}
with $\alpha$, $\beta$, $\gamma \in [0, 1]$ being the respective weights.

\paragraph{\model{} Inference.} The GCN module is only used during the distillation process. Hence, at test time, the student model performs zero-shot inference given just the question and uses only the same number of parameters and architecture as the base student model.

\begin{table*}[t]
\small
    \centering
    \vspace{-0.5em}
    \caption{Comparison of structured distillation (\model{}) with no teacher, single-teacher, and multi-teacher distillation baselines. Firstly, \model{} outperforms \emph{all} baselines across \emph{all} five reasoning benchmarks. On average, \model{} outperforms the strongest \textsc{SiT}-GPT4 baseline by 4.61\% and the no teacher baseline by 10.71\%. Secondly, knowledge distillation from each component of \graph{} improves the student model, as demonstrated by a consistent increase in performance from Level 1 to Level 4. Lastly, we also make all nodes available for a multi-teacher baseline, DSS-MT. \method{} obtains a larger improvement over \textsc{SiT}-GPT4 than DSS-MT does (2.49\% vs. 4.61\%).}
    \vspace{0.1in}
    \resizebox{\textwidth}{!}{%
    \begin{tabular}{lllccccc|c}
        \toprule
        \multirow{3}*{\textbf{Distillation Type}} & \multirow{3}*{\textbf{Distillation Data}} & \multirow{3}*{\textbf{Distilled Model}} & \multicolumn{5}{c}{\textbf{Datasets}} & \multirow{3}*{\textbf{Average Acc}} \\ \cmidrule{4-8}
        & & & \textbf{StrategyQA} & \textbf{CSQA} & \textbf{ARC-c} & \textbf{GSM8K} & \textbf{MATH} \\ \midrule
        No Teacher~\citep{jiang2023mistral} & - & Mistral-7B-Instruct & 61.57 & 57.89 & 60.32 & 44.05 & 7.02 & 46.17 \\ \midrule
         \rowcolor{gray!15} & Claude2 & \textsc{SiT}-Claude2 & 64.39 & 64.18 & 68.24 & 45.34 & 7.24 & 49.89 \\
         \rowcolor{gray!15} & Bard & \textsc{SiT}-Bard & 68.56 & 65.06 & 66.87 & 45.61 & 7.06 & 50.63 \\
          \rowcolor{gray!15} \multirow{-3}{4cm}{Single-Teacher~\citep{li-etal-2023-symbolic, magister2022teaching, fu2023specializing, ho2022large}} & GPT-4 & \textsc{SiT}-GPT4 & 69.96 & 66.87 & 68.91 & 47.38 & 8.24 & 52.27 \\
         \midrule
        \rowcolor{blue!10} & All Nodes & DSS-MT~\citep{hsieh2023distilling} & 71.18 & 69.42 & 71.38 & 51.84 & 9.98 & 54.76 [\textcolor{color3}{+ 2.49\%}]\\
         \rowcolor{blue!10} & Round-0 Nodes & \model{}-R0 [Level 1] & 71.18 & 67.36 & 72.06 & 48.52 & 9.72 & 53.77 [\textcolor{color2}{+ 1.50\%}]\\
         \rowcolor{blue!10} & Correct Nodes & \model{}-CN [Level 2] & 71.62 & 69.31 & 72.34 & 50.11 & 10.66 & 54.81 [\textcolor{color2}{+ 2.54\%}]\\
         \rowcolor{blue!10} & All Nodes & \model{}-AN [Level 3] & 72.10 & 70.65 & 71.92 & 50.69 & 11.98 & 55.47 [\textcolor{color2}{+ 3.20\%}]\\
         \rowcolor{blue!10} \multirow{-5}{2cm}{Multi-Teacher} & \graph{} & \model{} [Level 4] & \bf 74.24 & \bf 72.56 & \bf 72.61 &\bf 52.27 & \bf 12.76 & \bf 56.88 [\textcolor{color2}{+ 4.61\%}] \\
        \bottomrule
    \end{tabular}
    \vspace{-1em}
    }
    \label{table:main}
\end{table*}

\section{Experimental Setup}

\textbf{Implementation Details.} We test our structured distillation method, \model{}, with three instruction-tuned student models that are of different scales and belong to different model families: Mistral-7B-Instruct, LLaMA-2-7B-Chat, and LLaMA-2-13B-Chat. 
We train both task-specific distilled models (i.e., trained and tested on a single downstream task) and also one joint multi-task distilled model (trained on all in-domain tasks together and then tested on each in-domain and out-of-domain task). The multi-task model represents our unified model for OOD tasks. For conciseness, we refer \model{} to the \emph{resultant task-specific models} and \model{}-MT will refer to the \emph{resultant multi-task model}. See Appendix~\ref{appendix: implementation} for other implementation details.
\paragraph{Baselines.}
We group all methods into three categories of `Distillation Source Type', described as follows.
    \sparagraph{(1) No Teacher.} The lower bound of our distilled models is the zero-shot base model (e.g., Mistral-7B-Instruct).
    \sparagraph{(2) Single-Teacher.} Our next set of baselines only uses training data from a single teacher (out of the three agents used to construct \graph{}s). This follows multiple prior works~\citep{li-etal-2023-symbolic, magister2022teaching, fu2023specializing, ho2022large} that fine-tune student models on CoT reasoning using the next-token prediction objective (Equation~\ref{eqn:pos}). In particular, we fine-tune three student models with one of GPT-4, Bard, or Claude2 as the teacher using \emph{only} the positive samples from the respective teacher model. We will refer to these \textbf{Si}ngle-\textbf{T}eacher distilled models as \textsc{SiT}. Then \textsc{SiT}-GPT4, for example, will refer to a distilled model trained with only GPT4 as the teacher.
    \sparagraph{(3) Multi-Teacher.} Due to the lack of existing multi-teacher baselines, we first adapt an existing single-teacher method, Distilling Step-by-Step~\citep{hsieh2023distilling} to a multi-teacher setup. The other multi-teacher baselines (\method{}-*) correspond to distilled models trained with increasing levels of MAGs; these baselines demonstrate the utility of the levels as defined in Fig.~\ref{fig:intro}.
    \begin{itemize}[itemsep=1pt, wide=0pt, leftmargin=*, after=\strut]
    \item \textbf{DSS-MT.} ~\citet{hsieh2023distilling} propose `Distilling Step-by-Step (DSS)' with a multi-task objective to predict the label and rationale separately. We apply the same multi-task objective to all teachers to directly compare the effectiveness of the DSS objectives on the same MAG data. We refer to this as DSS-MT (DSS with Multi-Teacher).
    \item \textbf{\model{}-R0 (Level 1).} \model{}-R0 refers to a model that is fine-tuned only on the \emph{Round-0} (pre-interaction) correct reasoning of multiple teachers (i.e., GPT-4, Bard, and Claude2) using only the $\mathcal{L}^+$ objective defined in Eqn.~\ref{eqn:pos}, i.e., Level 1 of \method{}.
    
    \item \textbf{\model{}-CN (Level 2).} Next, \model{}-CN is a distilled model that is trained on \emph{all correct} (pre- and post-interaction) reasoning from all teachers with again the same $\mathcal{L}^+$ loss, corresponding to Level 2 of \method{}.

    \item \textbf{\model{}-AN (Level 3).} Going one step further, \model{}-AN is a distilled model that is trained on \emph{all nodes} (correct and incorrect) from all teachers. Hence, this is trained with both $\mathcal{L}^+$ and $\mathcal{L}^-$ objectives in Equations~\ref{eqn:pos} and~\ref{eqn:neg}. 
    Note that all three models for Levels 1-3 are \emph{unstructured} multi-teacher distilled models that view \graph{}s as a set of correct and incorrect nodes. 

    \item \textbf{\model{} (Level 4).} This is our final full method that distills knowledge from \emph{all nodes and edges} of a \graph{} using all three objectives as described in Equation~\ref{eqn:all}.

\end{itemize}

\vspace{-1.5em}
\section{Results and Analysis}
\subsection{Main Results}
\label{sec:main_results}
Our primary results demonstrate the effectiveness of \model{} across five reasoning benchmarks over different single-teacher and multi-teacher distillation setups. For the main results, we use Mistral-7B-Instruct as the student model and train task-specific distilled models (see Sec.~\ref{sec:scaling} for experiments with multi-task models with larger students belonging to different model families). We report accuracy for each task. Based on Table~\ref{table:main} results, we summarize our main conclusions below, addressing our research question posed in \cref{sec:intro}: How can we distill diverse teacher interactions into a smaller and more efficient student, utilizing the \emph{increasing levels of interaction structure in a MAG}? 

\sparagraph{Level 1: Distillation from multiple teachers outperforms distillation from the single strongest teacher.} Knowledge distillation from the correct reasoning chains of multiple teachers i.e., GPT-4, Bard, and Claude2, outperforms distillation from the single strongest teacher by an average of 1.50\% (see \model{}-R0 row versus \textsc{SiT}-GPT4 row). Different teachers bring diversity in their reasoning, leading to improved reasoning capabilities of the student model. 

\sparagraph{Level 2: Distillation from pre- and post-interaction reasoning outperforms only pre-interaction reasoning.} Training a student model on all correct reasoning chains after multiple teacher models have interacted further improves the student's reasoning. Fine-tuning on post-interaction reasoning chains effectively increases the training data and its usefulness is validated through an overall 2.54\% improvement, compared to the best single-teacher model (see \model{}-CN row versus \textsc{SiT}-GPT4 row).

\sparagraph{Level 3: Negative reasoning chains help.} Additionally, learning by contrasting between positive and negative chains improves over all previous levels. Our \model{}-AN model obtains an overall average improvement of 3.20\% (see \model{}-AN row versus \textsc{SiT}-GPT4 row).

\sparagraph{Level 4: Structured distillation from interactions outperforms all multi-teacher baselines.} Our final structured distillation method, \model{}, that distills knowledge from both the outputs (nodes) and structure (edges) of multi-agent interactions, obtains the largest improvement and outperforms all single- and multi-teacher models. When applied to Mistral-7B-Instruct,
\model{} surpasses a model that learns only from GPT4 by an average of 4.61\%. 
Out of all the levels, distillation from interactions brings the most improvements (both individually and across all five benchmarks), specifically demonstrating the utility of modeling the conversation structure of multi-agent discussions. Overall, compared to the base model, \model{} improves the student's reasoning capabilities by a significant 10.71\% (46.17 $\rightarrow$ 56.88) while maintaining similar inference-time efficiency. 

\sparagraph{\method{} also outperforms adapted single-teacher distillation from prior work.} When using data from all teacher models, \method{} consistently outperforms DSS-MT. Since the learning source is the same in both methods (i.e., all nodes in MAGs), it underscores the effectiveness of learning from teacher interactions and our proposed objectives.

\sparagraph{\model{} improves inference efficiency compared to \textsc{ReConcile}.} Compared to \textsc{ReConcile}, \model{} also drastically improves inference efficiency.
Because the sizes of gated models like those used in \textsc{ReConcile} are not known, we measure efficiency via number of output tokens generated.\footnote{This metric is extremely strict and under-estimates our efficiency gains, as each of the gated LLMs used by \textsc{ReConcile} exceeds 7B parameters, and each example for \textsc{ReConcile} involves multiple inference calls.}
As shown in Table~\ref{tab:token_count}, \model{} achieves up to a $9$x reduction in token count. 
\model{} obtains this efficiency by always answering questions in one inference call; this differs from \textsc{ReConcile} which has to go through up to twelve (expensive) LLM inference calls (involving 3 agents for the initial round and another 3 rounds).
Moreover, \cref{fig:pareto} and Table~\ref{tab:pareto} in Appendix shows the trade-off between token efficiency and performance, averaged across datasets and for each individual dataset respectively.
While \textsc{ReConcile} (as the upper-bound) has the best performance, it also produces the most tokens (cf. \cref{tab:token_count}); the zero-shot and single-teacher models are more efficient than \textsc{ReConcile} but suffer in terms of performance. 
On the other hand, \method{} achieves a better balance of efficiency and performance than these baselines, with more efficiency than \textsc{ReConcile} and higher performance than the zero-shot and prior single-teacher distillation methods. 

\begin{figure}[t]
    \centering
    \includegraphics[width=\linewidth]{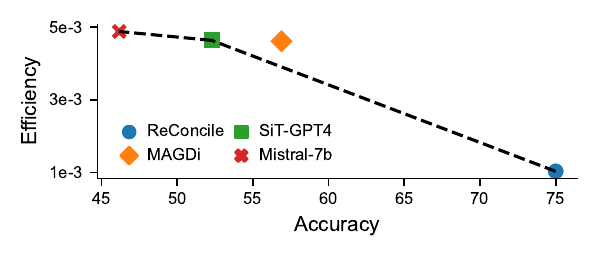}
    \vspace{-2.5em}
    \caption{Trade-off between performance and efficiency. \method{} exceeds the Pareto frontier of prior work, surpassing single-teacher models in performance and surpassing \textsc{ReConcile} in efficiency, defined as $1/avg(tokens)$.}
    \vspace{-1.5em}
    \label{fig:pareto}
\end{figure}

\begin{table}[]
\centering
\caption{Comparison of the token counts generated by \textsc{ReConcile} (a multi-agent interaction framework) and \model{}.}
    \vspace{0.1in}
\resizebox{0.8\columnwidth}{!}{%
\begin{tabular}{lccc}
\toprule
           & \textbf{\textsc{ReConcile}} & \textbf{\model{}} & \textbf{Reduction} \\\midrule
StrategyQA & 924.5     & 107.5   & 8.6x \\
CSQA       & 936.9     & 104.2   & 9.0x\\
ARC-c      & 448.3     & 86.4    & 5.2x\\
GSM8K      & 642.3     & 141.6   & 4.5x\\
MATH       & 1900.1    & 645.0   & 2.9x\\
\midrule
Average    & 970.4     & 216.9   & 4.5x\\
\bottomrule
\end{tabular}
}
\label{tab:token_count}
\vspace{-1.5em}
\end{table}

\sparagraph{\method{} effectively transfers multi-agent capabilities into a single student.}
In Table~\ref{tab:transfer}, we compare distilled student models to their teachers by reporting results for (1) GPT4, as the upper-bound performance of a single-teacher, (2) \textsc{ReConcile}, as the upper-bound of multi-agent teacher (using GPT4, Claude2, Bard), (3) SiT-GPT4, as distillation from the single-teacher, and (4) \method{} as distillation from multi-agent teacher. We also report the relative improvement from `single' to `multi', both with and without distillation. 
The relative improvement from SiT-GPT4 to \method{} (12.70\%) is much higher than that from GPT-4 to the multi-teacher system \textsc{ReConcile} (5.79\%), highlighting the effectiveness of \method{} in transferring multi-agent capabilities into a single student model.

\subsection{Analysis: Generalizability, Scalability, Diversity}
\label{sec:scaling}

\sparagraph{\model{} can be used to train one joint multi-task model.} In our main experiments (Sec.~\ref{sec:main_results}), we fine-tuned task-specific models with structured distillation. While these multi-teacher task-specific models show clear benefits over single-teacher models, we would ideally like \emph{one joint} model that can tackle all tasks. Therefore, we train a joint multi-task model (\model{}-MT) by combining training data from all five benchmarks and evaluating it together on all benchmarks. \model{}-MT obtains an average accuracy that is within 1\% of task-specific models (56.89\% versus 55.12\%), showing its applicability for training a joint model (refer to Appendix Table~\ref{tab:multitask} for the full table).

\begin{table}[t]
\centering
\caption{Comparison of single-teacher and multi-teacher distilled students with their respective teachers. The relative improvement from SiT-GPT4 to \method{} (12.70\%) is higher than that from GPT-4 to the multi-teacher system ReConcile (5.79\%).}
\vspace{0.1in}
\resizebox{0.95\columnwidth}{!}{%
\begin{tabular}{lcccccc}
\toprule
& \textbf{StrategyQA} & \textbf{CSQA} & \textbf{ARC-c} & \textbf{GSM8K} & \textbf{MATH} & \textbf{Avg}   \\\midrule
GPT4       & 75.60                & 73.30          & 94.50           & 90.70           & 39.00          & 74.62          \\
ReConcile  & 87.70                & 78.70          & 96.30           & 92.10           & 41.00          & 79.16          \\
\% improved & 13.80               & 6.86          & 1.87           & 1.52           & 4.88          & \textbf{5.79}  \\\midrule
SiT-GPT4   & 69.96               & 66.87         & 68.91          & 47.38          & 8.24          & 52.27          \\
\method{}      & 74.24               & 72.56         & 72.61          & 52.27          & 12.76         & 56.88          \\
\% improved & 5.77                & 7.84          & 5.10           & 9.36           & 35.42         & \textbf{12.70}\\\bottomrule
\end{tabular}}
\label{tab:transfer}
\end{table}

\begin{table}[t]
\small
\centering
\vspace{-1em}
\caption{Out-of-domain comparison between Single-Teacher Multi-Task (\textsc{SiT}-GPT4-MT) and \model{} Multi-Task (\model{}-MT) models. \model{}-MT performs up to 7\% better than the single-teacher baseline even on OOD datasets (57.52 vs. 64.30).
}
    \vspace{0.1in}
\begin{tabular}{lccc}
\toprule
          &    \textbf{BoolQ}           & \textbf{SVAMP} \\\midrule
SiT-GPT4-MT & 60.70      & 57.52          \\
\method{}-MT  &    \textbf{63.98}           & \textbf{64.30 }         \\\bottomrule
\end{tabular}
\vspace{-1.5em}

\label{tab:ood}
\end{table}

\sparagraph{\model{} generalizes to OOD tasks.} We also evaluate \model{}-MT on two out-of-domain benchmarks (BoolQ for commonsense reasoning and SVAMP for math) that were not included in multi-task training. As shown in Table~\ref{tab:ood}, \model{} outperforms single-teacher distillation by to $3\%$ on BoolQ and $7\%$ on SVAMP. The broader implication of this is that the single takeaway \model{}-MT model maintains good performance on OOD tasks and continues to outperform single-teacher baselines on new datasets.

\begin{figure}[t]
    \centering
    \vspace{0.5em}
    \includegraphics[width=0.8\linewidth]{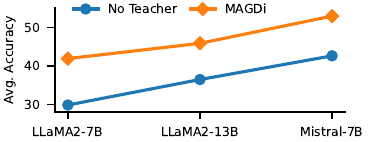}
    \vspace{-1em}
    \caption{Scaling results of \model{} with different base student models. As the average (zero-shot) performance of the base model improves (Mistral-7B $>$ LLaMA-2-13B$>$ LLaMA-2-7B), \model{} shows a corresponding increase.
    }
    \vspace{-1.4em}
    \label{fig:scaling_average}
\end{figure}

\sparagraph{\model{} scales positively with better base models.} We now study the scaling properties of structured distillation by varying the base student model. In particular, we train three structurally distilled models with \model{}, using LLaMA-2-7B-Chat, LLaMA-2-13B-Chat, and Mistral-7B-Instruct as the base models. In Fig.~\ref{fig:scaling_average}, we plot the average accuracy of all five benchmarks (and see Table~\ref{tab:architecture} for individual results). The ordering of the three models is based on their zero-shot performance. We observe that \model{} brings consistent improvements over all base models, maintaining proportionate gains. Overall, the scaling trend of \model{} suggests that structured distillation should continue to improve even stronger students. See Appendix~\ref{appendix:analysis} for other analyses of \model{} e.g., the effect of the amount of training data and different graph structures.

\sparagraph{\model{} boosts self-consistency.} We hypothesize that a student model that learns from \emph{multiple teachers} will exhibit better diversity in its generations. To test this, we combine \model{} with an ensemble method like Self-Consistency (SC)~\cite{wang.y.2022self}. SC computes a majority vote over model answers and has proven effective for reasoning tasks; SC's improvements are predicated on answer diversity (since a majority vote over the versions of the same answer would not yield any improvements). We show that self-consistency with our \emph{multi-teacher distilled model} outperforms the same with (1) the base model and more importantly, (2) a single-teacher distilled model. As shown in Table~\ref{tab:sc}, on GSM8K, SC between 10 responses improves accuracy by 15\% when applied to a \model{}-trained student, compared to only 4\% improvement for the base model (Mistral-7B-Instruct) and 11\% for the single-teacher model (\textsc{SiT}-GPT4). Broadly, this suggests that inference-time algorithms relying on a model's inherent diversity can be boosted from multi-teacher distillation.

\sparagraph{GCN outperforms self-attention for modeling MAGs.}
\looseness-1
To demonstrate the effectiveness of the GCN in encoding MAGs, we compare it to two alternative approaches. 
These remove the explicitly-defined graph structure of the GCN (which is determined by the MAG) and instead aim to automatically learn the interactions with multi-head attention layers~\citep{vaswani2017attention}.
\emph{Token-level attention:} Here we linearize the entire interaction (across agents and rounds) into a single sequence of tokens, so that any token (across all reasoning chains, rounds, and agents) can attend to any other token.
\emph{Node-level attention:} Here we keep the nodes intact, i.e., each node still represents a whole reasoning chain, but we linearize them into a sequence of nodes so that any node can attend to any other node.
Results presented in Table~\ref{tab:structure_ablation} show that visualizing the entire interaction as a sequence of tokens results in long token lengths and removes the boundaries between the nodes (reasoning chains), leading to significantly worse performance (first row). Node-level attention improves slightly over token-level attention by keeping the nodes (chains) intact and modeling the interactions between them with self-attention layers (second row).
However, our GCN-based approach performs the best across all datasets. While the transformer could theoretically model any graph structure, it may also require a large amount of training data and tuning to infer the correct interaction graph structure. Hence, we find that defining the interaction patterns with MAGs apriori and modeling them with a GCN is more performant and data-efficient. Note that \method{}’s main contributions are in defining structured representations of multi-agent interactions and then distilling such interaction data into a weaker student model. This contribution is independent of the exact graph representation learning module, and going forward, we hope that our results will motivate newer methods of modeling these multi-agent interactions.

\begin{table}[t]
\centering
\small
\caption{Self-consistency with \model{} on GSM8K achieves the largest gain (up to 15\%) compared to the same with the base student model and the single-teacher distilled model.}
\vspace{0.1in}
    \resizebox{\columnwidth}{!}{%
\begin{tabular}{lccc}
\toprule
          &    \textbf{Mistral-7B}           & \textbf{\textsc{SiT}-GPT4} & \textbf{\model{}} \\\midrule
w/o SC & 44.05      & 47.38    & 52.27      \\
w/ SC &  48.44 [\textcolor{color2}{+ 4.39\%}] & 58.62 [\textcolor{color2}{+ 11.24\%}] & \bf 67.42 [\textcolor{color2}{+ \textbf{15.15\%}}]       \\\bottomrule
\vspace{-3em}
\end{tabular}
}

\label{tab:sc}
\end{table}

\begin{table}[t]
\caption{Comparison of different graph modeling methods. \method{} with GCN consistently outperforms attention-based variants for modeling interaction graphs.}
 \vspace{0.1in}
\resizebox{\columnwidth}{!}{%
\begin{tabular}{lccccc}
\toprule
\textbf{Method} & \textbf{StrategyQA} & \textbf{CSQA}  & \textbf{ARC-c} & \textbf{GSM8K} & \textbf{MATH}  \\ \midrule
Token Attn      & 68.56               & 65.04          & 70.31          & 51.17          & 8.86           \\
Node Attn       & 69.87               & 71.16          & 70.01          & 51.63          & 9.22           \\
GCN             & \textbf{74.24}      & \textbf{72.56} & \textbf{72.61} & \textbf{52.27} & \textbf{12.76} \\ \bottomrule
\label{tab:structure_ablation}
\end{tabular}}
\vspace{-2.5em}
\end{table}

\section{Discussion and Conclusion}

We have tackled the problem of structured distillation from multi-agent interactions as a way to equip much smaller and more efficient language models with improved reasoning capabilities. To achieve this goal, we proposed a graph-based representation of these interactions, generated graphs for training, and developed a structured distillation method for learning from these interaction graphs. Our results showed the effectiveness, generalizability, and scalability of structured distillation across multiple reasoning benchmarks.

\looseness-1
While \method{} relies on interactions between LLMs for supervision, its modular design makes it well-suited even to scenarios where such data may be limited. 
Revisiting \method{}’s four objectives, we split its use cases into the following four categories. (1) \emph{No teacher data is available:} When no teacher data is available, we demonstrate \method{}'s ability to generalize in a zero-shot manner (Table~\ref{tab:ood}). (2) \emph{Only single-teacher data is available:} When supervision from only one teacher model is available, we can leverage the first objective in \method{} (Level 1). (3) \emph{Having multi-teacher data but no interaction:} When multiple teacher sources are present but the interactions are absent, we show gains from the first two \method{} objectives (Level 3). (4) \emph{Having multi-teacher interaction data:} This is when the full \method{} model (Level 4) can be applied to maximize performance. Thus, our four levels succinctly depict the adaptability of \method{} based on the amount and nature of data availability, providing a structured framework for implementing its objectives across varying data scenarios and applications.

\section*{Acknowledgements}

We thank Peter Hase, Archiki Prasad, and the anonymous reviewers for useful feedback and suggestions regarding experiments. This work was supported by NSF-CAREER Award 1846185, NSF-AI Engage Institute DRL-2112635, DARPA MCS Grant N66001-19-2-4031, Accelerate Foundation Models Research program, and a Google PhD Fellowship. The views contained in this article are those of the authors and not of the funding agency.

\section*{Impact Statement}
The computing resources used by LLMs incur a substantial carbon footprint \citep{strubell2019energy}, and running them in interactive multi-agent settings like \textsc{ReConcile} further increases these environmental costs. 
\method{} distills from several LLMs into a single more efficient and smaller LM, thus reducing the environmental cost of running LLMs while still maintaining strong performance. 

The LLMs \method{} distills from -- and the student models it distills into -- can reflect stereotypes, biases, and other negative traits present in their pre-training data \citep{weidinger2021ethical}, which we do not have control over. 
Our distilled LLMs have the same capacity for undesirable generation as their teacher models and their respective zero-shot variants; as such, the models resulting from \method{} distillation have the same potential for misuse as any LLM or method distilling from LLMs. 
More studies are needed to evaluate and mitigate such biases in LLMs.

\bibliography{custom}
\bibliographystyle{icml2024}

\appendix

\section*{Appendix}
\begin{figure}[h]
    \centering
    \includegraphics[width=0.4\textwidth]{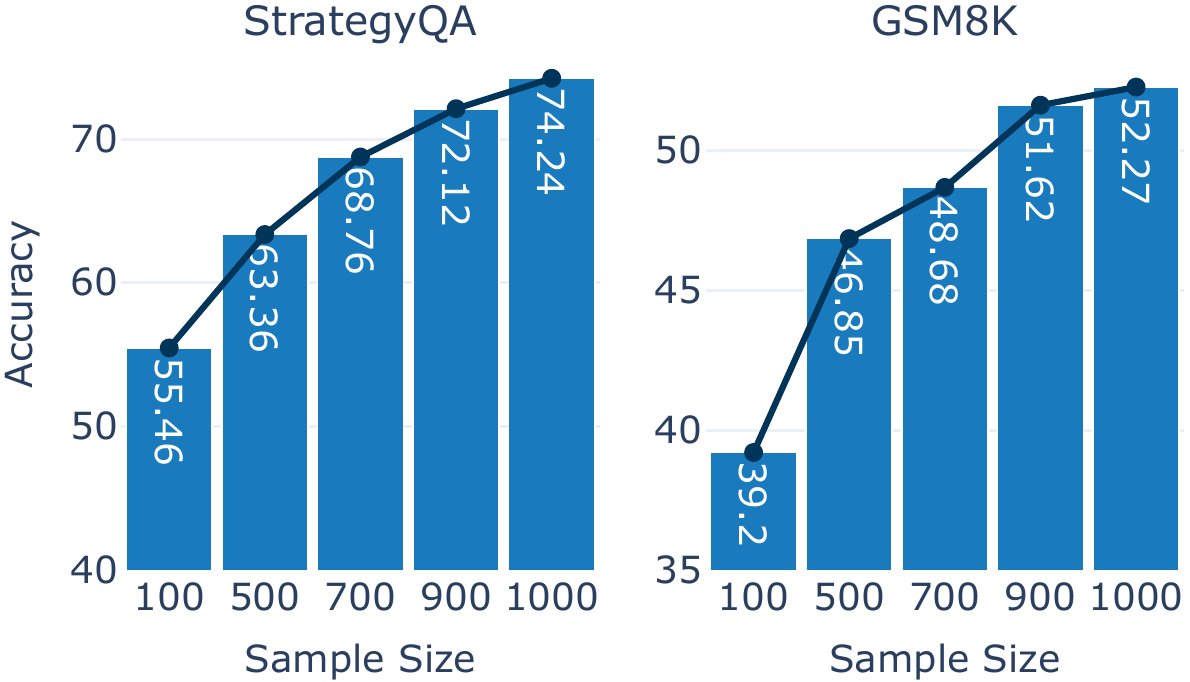}
    \caption{Results with scaling training data on StrategyQA and GSM8K. With structured distillation, student accuracy increases with an increase in training data.}
    \label{fig:data_scaling}
\end{figure}

\section{Implementation Details of \model{}}
\label{appendix: implementation}
\paragraph{Learning from Negative Reasoning.} We provide additional details about \model{}'s contrastive objective in Equation~\ref{eqn:neg}. In practice, since a \graph{} can have many nodes, computing the contrastive loss for every possible positive and negative pair can be prohibitively expensive. In such cases, we randomly sample nodes from the minority group to pair up with the nodes from the majority group. For instance, if there are 5 positive and 3 negative nodes in a \graph{}, then it naturally forms three pairs and for the remaining 2 positives, we randomly sample 2 negatives to pair with them.
\paragraph{Model Training.} \model{}-distilled models are fine-tuned using Low-Rank Adaptation (LoRA) for efficiency~\cite{hu2022lora}. We set the rank to 16 and alpha to 32. We fine-tune the student model for 10 epochs using a learning rate of 5e-6 and a batch size of 16. For the hyperparameters $\alpha$, $\beta$ and $\gamma$, please refer to our code implementation for detailed settings in each dataset. All of our experiments are run on four RTX A6000 with 48G memory each. 

\section{Further Analyses of \model{}}
\label{appendix:analysis}
\sparagraph{\model{} scales positively with better base models.} In Table~\ref{tab:architecture}, we show \model{}'s scaling trends on StrategyQA, CommonsenseQA, GSM8K and MATH. Here we apply \model{} to three base models: Mistral-7B-Instruct, LLaMA-2-7B-Chat, and LLaMA-2-13B-Chat. Across all these benchmarks, \model{} demonstrates consistent gains on top of all base models.

\begin{table}[]
\small
\centering
\caption{Results of \model{} with different base student models (LLaMA-2-7B-Chat, LLaMA-2-13B-Chat, and Mistral-7B-Instruct). First, across all benchmarks and for all three base models, \model{} outperforms no teacher baselines by a large margin. Second, the effect of structured distillation correlates with the performance of the base model (Mistral-7B $>$ LLaMA-2-13B $>$ LLaMA-2-7B), highlighting the scaling properties of \model{}.}
\vspace{0.1in}
\resizebox{\columnwidth}{!}{%
\begin{tabular}{llcccc}
\toprule
\textbf{Base} & \textbf{Distilled} & \textbf{StrategyQA} & \textbf{CSQA} & \textbf{GSM8K} & \textbf{MATH} \\\midrule
\multirow{2}{*}{LLaMA-2-7B}               & No Teacher                                   & 51.53                                   & 46.81                             & 18.60                              & 2.50                              \\
                                        & MAGDi                                        & \textbf{66.81}                          & \textbf{66.73}                    & \textbf{28.36}                     & \textbf{5.76}                     \\\midrule
\multirow{2}{*}{LLaMA-2-13B}             & No Teacher                                   & 58.52                                   & 51.73                             & 31.77                              & 3.90                              \\
                                        & MAGDi                                        & \textbf{69.00}                          & \textbf{67.05}                    & \textbf{40.56}                     & \textbf{6.46}                     \\\midrule
\multirow{2}{*}{Mistral-7B}             & No Teacher                                   & 61.57                                   & 57.89                             & 44.05                              & 7.02                              \\
                                        & MAGDi                                        & \textbf{74.24}                          & \textbf{72.56}                    & \textbf{52.27}                     & \textbf{12.76}                   
\\\bottomrule
\end{tabular}
    }
\label{tab:architecture}
\end{table}

\begin{table}[]
\small
\centering
\caption{Dataset licenses}
\vspace{0.1in}
\begin{tabular}{ll}
\toprule
\textbf{Dataset} & \textbf{License} \\ \midrule
StrategyQA & MIT License (\href{https://github.com/eladsegal/strategyqa/blob/main/LICENSE}{License}) \\ 
CommonsenseQA & MIT License (\href{https://github.com/jonathanherzig/commonsenseqa/issues/5}{License}) \\ 
ARC-c & CC BY-SA 4.0 (\href{https://allenai.org/data/arc}{License}) \\ 
GSM8K & MIT License (\href{https://github.com/openai/grade-school-math/blob/master/LICENSE}{License}) \\ 
MATH & MIT License (\href{https://github.com/hendrycks/math/blob/main/LICENSE}{License}) \\ 
BooQ & CC BY-SA 3.0 (\href{https://github.com/google-research-datasets/boolean-questions}{License}) \\ 
SVAMP & MIT License (\href{https://github.com/arkilpatel/SVAMP/blob/main/LICENSE}{License}) \\
\bottomrule
\end{tabular}
\label{tab:dataset_licenses}
\end{table}

\vspace{1em}
\sparagraph{\model{} scales positively with the amount of training data.} Next, we analyze the scaling properties of \model{} by varying the amount of training data. We train distilled models with \model{} (using Mistral-7B-Instruct as the base model) by varying the amount of training data from 100 to 1000 samples. Figure~\ref{fig:data_scaling} shows that with more data, \model{} exhibits better performance -- e.g., on StrategyQA, training on 1K \graph{}s improves reasoning performance by 10.88\% compared to training on 500 samples. This suggests that \model{} may bring additional improvements with a larger training corpus.

\sparagraph{Dense interaction graphs distill significant knowledge to students.} Recall that the \graph{}s in our corpus have distinct structures, with $\mathcal{G}_3$ being the densest graph (having the most nodes and edges). We show that removing these $\mathcal{G}_3$ graphs from our training corpus leads to a significant drop in student accuracy. As shown in Table~\ref{tab:structure_analysis}, structured distillation on CSQA without the $\mathcal{G}_3$ graphs causes student performance to drop by 2\% and additionally removing the $\mathcal{G}_2$ graphs causes a further drop of 1\%. Our result thus highlights the importance of distillation from denser graphs (even if they are sparsely represented in the corpus).

\sparagraph{MAG edges effectively model the discussion structure.} 
Recall that we defined directed edges in MAGs according to the interaction pattern between agents across rounds. To further demonstrate the utility of these edges, we compare our directed MAGs (D-MAG) to fully-connected MAGs (FC-MAG) where every node is connected to every other node -- and undirected MAGs (UD-MAG) where the edges follow the interaction but are undirected. As shown in Table~\ref{tab:graph_ablation}, FC-MAG performs significantly worse, showing the utility of defining edges according to interaction patterns. Finally, the directionality of the edges has marginal impact on the final performance.
\begin{table}[h!]
\small
\centering
\caption{Removing denser interaction graphs (e.g., $\mathcal{G}_3$ or $\mathcal{G}_2$) from CSQA training corpus leads to a significant drop in student accuracy, highlighting the importance of learning from such graphs.}
\label{tab:structure_analysis}
    \vspace{0.1in}
\begin{tabular}{lc}
\toprule
    \textbf{Training \graph{}s}         & \textbf{Accuracy} \\\midrule
All          & 72.56             \\
w/o $\mathcal{G}_3$   & 70.65             \\
w/o $\mathcal{G}_2$ \& $\mathcal{G}_3$ & 69.65             \\
\bottomrule
\end{tabular}
\end{table}

\begin{table}[h!]
\caption{Dataset-wise breakdown of the performance and efficiency trade-off shown in Fig~\ref{fig:pareto}.}
\vspace{0.1in}
\resizebox{\columnwidth}{!}{%
\begin{tabular}{lcccccc}
\toprule
& \textbf{StrategyQA} & \textbf{CSQA}  & \textbf{ARC-c} & \textbf{GSM8K} & \textbf{MATH}   & \textbf{Avg}   \\\midrule
\# Tok (ReConcile) & 924.5      & 936.9 & 448.3 & 642.3 & 1900.1 & 970.4 \\
\# Tok (MAGDi)     & 107.5      & 104.2 & 86.4  & 141.6 & 645.0  & 216.9 \\
Reduction            & 8.6x       & 9.0x  & 5.2x  & 4.5x  & 2.9x   & 4.5x  \\\midrule
Acc (ReConcile) & 79.0       & 74.7  & 93.5  & 85.3  & 41.0   & 74.7  \\
Acc (MAGDi)     & 74.2       & 72.6  & 72.6  & 52.3  & 12.8   & 56.9 \\\bottomrule
\end{tabular}}
\label{tab:pareto}
\end{table}

\begin{table}[h!]
\caption{Utility of defining edges according to interaction structure in MAGs. Fully-connected MAGs perform worse than defining (directed or undirected) edges based on interactions.}
 \vspace{0.1in}
\resizebox{\columnwidth}{!}{%
\begin{tabular}{lccccc}
\toprule
\textbf{Method} & \textbf{StrategyQA} & \textbf{CSQA}  & \textbf{ARC-c} & \textbf{GSM8K} & \textbf{MATH}  \\ \midrule
FC-MAG             & 71.59               & 71.22          & 69.38          & 50.29          & 11.02           \\
UD-MAG              & 74.02               & \textbf{72.61} & 72.46          & 52.10          & 12.66          \\
D-MAG              & \textbf{74.24}      & 72.56          & \textbf{72.61} & \textbf{52.27} & \textbf{12.76} \\
\bottomrule
\label{tab:graph_ablation}
\end{tabular}}
\end{table}
\begin{table*}[]
    \centering
    \small
    \caption{Training and test statistics for five benchmarks. Our training corpus (i.e., \graph{}s) is categorized along three dimensions: (1) \textbf{Round:} number of nodes belonging to each interaction round, (2) \textbf{Agent:} number of nodes belonging to each agent (GPT4/Bard/Claude2), and (3) \textbf{Graph:} number of graphs with a specific graph structure.}
    \vspace{0.1in}
    \begin{tabular}{lcccc}
    \toprule
     \multirow{3}{*}{\textbf{Task}} &  \multicolumn{3}{c}{\textbf{Train}} &  \multirow{3}{*}{\textbf{Test}} \\ \cmidrule{2-4}
     & \textbf{Round (0 / 1 / 2 / 3)} & \textbf{Agent (Each)} & \textbf{Graph ($\mathcal{G}_0$ / $\mathcal{G}_1$ / $\mathcal{G}_{2}$ / $\mathcal{G}_{3}$ / All)} \\
    \midrule
    StrategyQA & 3K / 843 / 438 / 102 & 1.4K & 719 / 135 / 112 / 34 / \textbf{1K} &   229 \\
    CSQA & 3K / 2.1K / 942 / 627 & 2.2K & 306 / 380 / 105 / 209 / \textbf{1K} &   2.2K \\
    ARC-c & 3K / 792 / 288 / 153 & 1.4K & 736 / 168 / 45 / 51 / \textbf{1K} & 1.1K \\
    GSM8K & 3K / 1.3K / 588 / 354 & 1.7K & 557 / 247 / 78 / 118 / \textbf{1K} &   1.3K \\
    MATH & 3K / 2.3K / 1.5K / 1.2K & 2.7K & 215 / 269 / 89 / 427 / \textbf{1K} &   5K \\
    \bottomrule
    \end{tabular}
    \label{tab:dataset}
\end{table*}

\begin{table*}[]
\small
    \centering
    \caption{Comparison between task-specific \model{} models and the joint multi-task \model{} model (\model{}-MT). On average, \model{}-MT's accuracy is within 1\% of task-specific models.}
    \vspace{0.1in}
    \begin{tabular}{lcccccc}
    \toprule
    \textbf{Distilled Model} & \textbf{StrategyQA} & \textbf{CSQA} & \textbf{ARC-c} & \textbf{GSM8K} & \textbf{MATH} & \textbf{Average Accuracy} \\ \midrule
        \model{}-MT & 70.74 & 71.73 & 72.87 & 49.51 & 10.76 & 55.12 \\
        \model{} & 74.24 & 72.56 & 72.61 & 52.27 & 12.76 & 56.89 \\
        \bottomrule
    \end{tabular}
    \label{tab:multitask}
\end{table*}

\newpage

\begin{figure*}[t]
    \centering
    \includegraphics[width=0.9\textwidth]{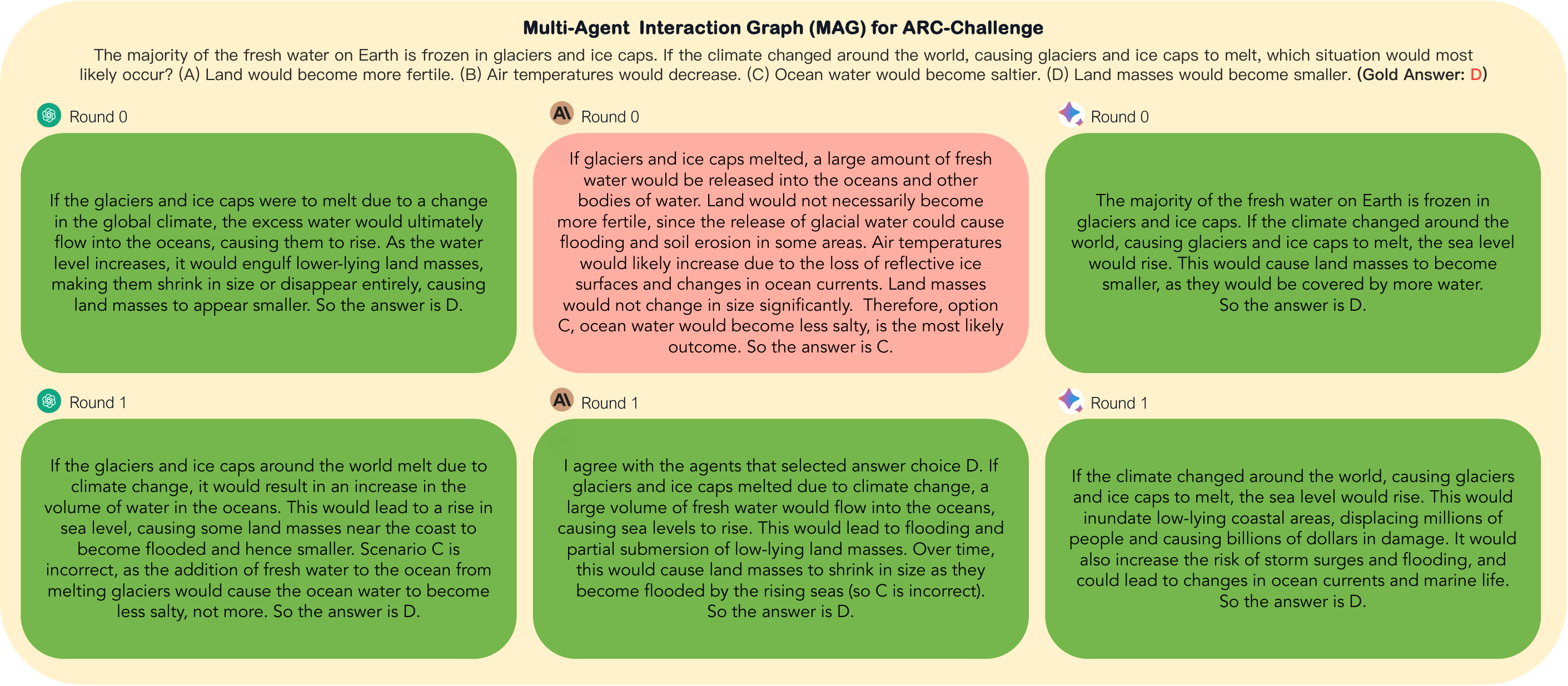}
    \caption{An example of a \graph{} from the ARC-c dataset.}
    \label{fig:MAG_ARC}
\end{figure*}
\begin{figure*}[t]
    \centering
    \includegraphics[width=0.9\textwidth]{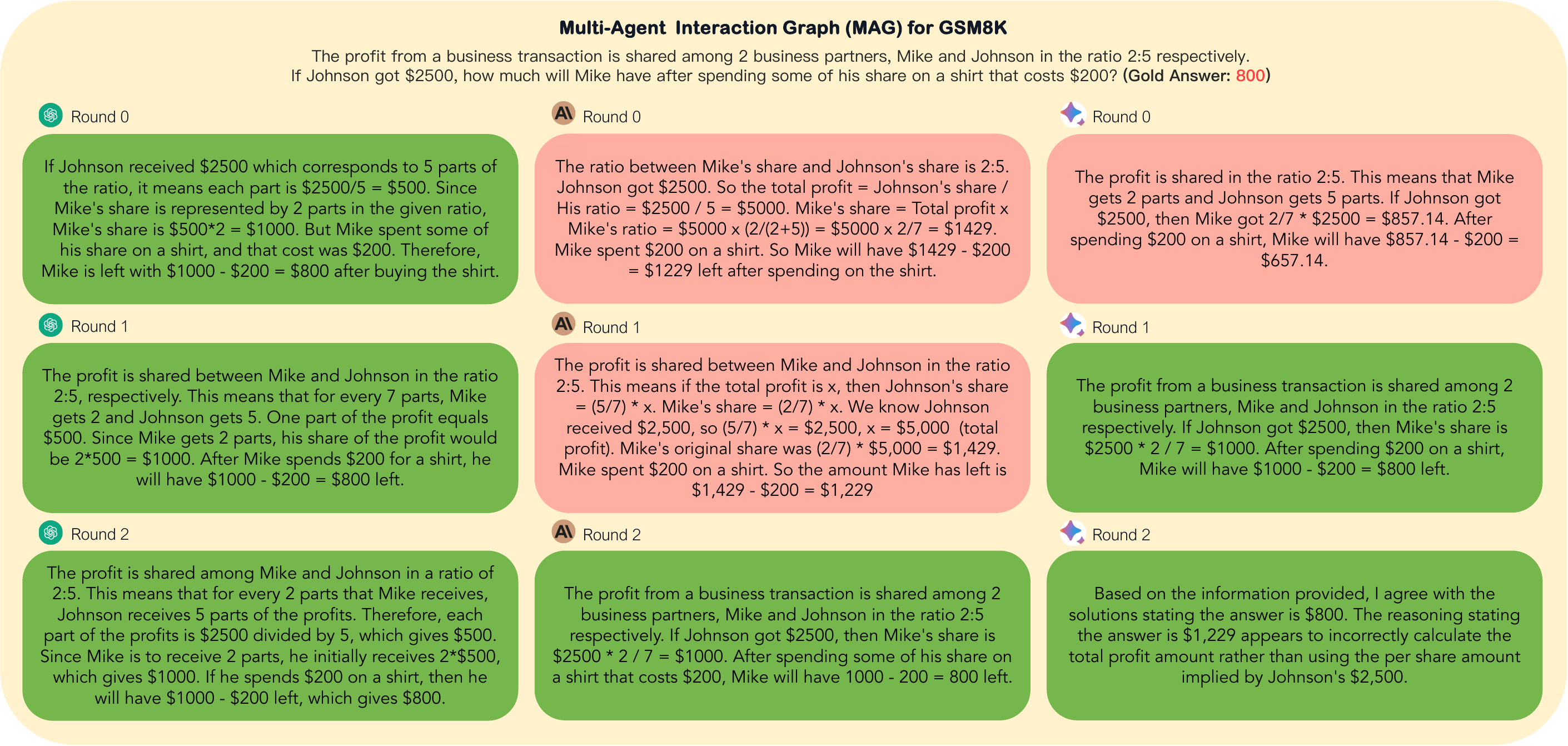}
    \caption{An example of a \graph{} from the GSM8K dataset.}
    \label{fig:MAG_GSM8K}
\end{figure*}

\section{Benchmark Licenses}
The licenses of the seven datasets we used are in Table~\ref{tab:dataset_licenses}.

\section{Qualitative Examples of \graph{}}
\label{appendix:qual}
See Fig.~\ref{fig:MAG_ARC} and Fig.~\ref{fig:MAG_GSM8K} for two examples of \graph{}s.
\end{document}